\documentclass[10pt,twocolumn,letterpaper]{article}

\usepackage[pagenumbers]{cvpr} 
\usepackage{multirow}
\usepackage[accsupp]{axessibility} 

\usepackage[dvipsnames]{xcolor}

\definecolor{cvprblue}{rgb}{0.21,0.49,0.74}
\usepackage[pagebackref,breaklinks,colorlinks,citecolor=cvprblue]{hyperref}

\def\paperID{800} 
\def\confName{CVPR}
\def\confYear{2024}

\title{LocLLM: Exploiting  Generalizable Human Keypoint Localization \\ via Large Language Model}

\author{Dongkai Wang \ \quad Shiyu Xuan \ \quad Shiliang Zhang\\ 
National Key Laboratory for Multimedia Information Processing,\\ School of Computer Science, Peking University\\
{\tt\small \{dongkai.wang, slzhang.jdl\}@pku.edu.cn, shiyu\_xuan@stu.pku.edu.cn}
}

\begin{document}
\maketitle
\begin{abstract}
The capacity of existing human keypoint localization models is limited by keypoint priors provided by the training data. To alleviate this restriction and pursue more general model, this work studies keypoint localization from a different perspective by reasoning locations based on keypiont clues in text descriptions. We propose LocLLM, the first Large-Language Model (LLM) based keypoint localization model that takes images and text instructions as inputs and outputs the desired keypoint coordinates.
LocLLM leverages the strong reasoning capability of LLM and clues of keypoint type, location, and relationship in textual descriptions for keypoint localization. To effectively tune LocLLM, we construct localization-based instruction conversations to connect keypoint description with corresponding coordinates in input image, and fine-tune the whole model in a parameter-efficient training pipeline. LocLLM shows remarkable performance on standard 2D/3D keypoint localization benchmarks. Moreover, incorporating language clues into the localization makes LocLLM show superior flexibility and generalizable capability in cross dataset keypoint localization, and even detecting novel type of keypoints unseen during training\footnote[2]{Project page: \url{https://github.com/kennethwdk/LocLLM}}.
\end{abstract}

\section{Introduction}
\label{sec:intro}

Human keypoint localization aims to locate target keypoints from input person image and is a fundamental task in computer vision and graphics. It has a wide range of applications in human pose estimation~\cite{sun2019deep,wang20233d,wang2023contextual,wang2023humvis} and facial landmark detection~\cite{sagonas2013300}, \emph{etc}. Existing keypoint localization methods typically utilize powerful neural networks, \emph{e.g.}, Convolutional Neural Network (CNN)~\cite{xiao2018simple,sun2019deep} or Vision Transformer (ViT)~\cite{xu2022vitpose} to either directly regress keypoint coordinates~\cite{li2021human} or estimate the keypoint heatmaps~\cite{xiao2018simple,sun2019deep} to perform localization. Those methods learn keypoint priors encoded in the training set into the backbone, \emph{e.g.}, along channel dimension in the last layer. This design reinforces the model responses on keypoints encoded in the backbone, but limits the generalization capability to detect keypoints in unseen human pose from different dataset, or to handle novel type of keypoint not included in the training set.

\begin{figure}[t]
  \centering
   \includegraphics[width=1.0\linewidth]{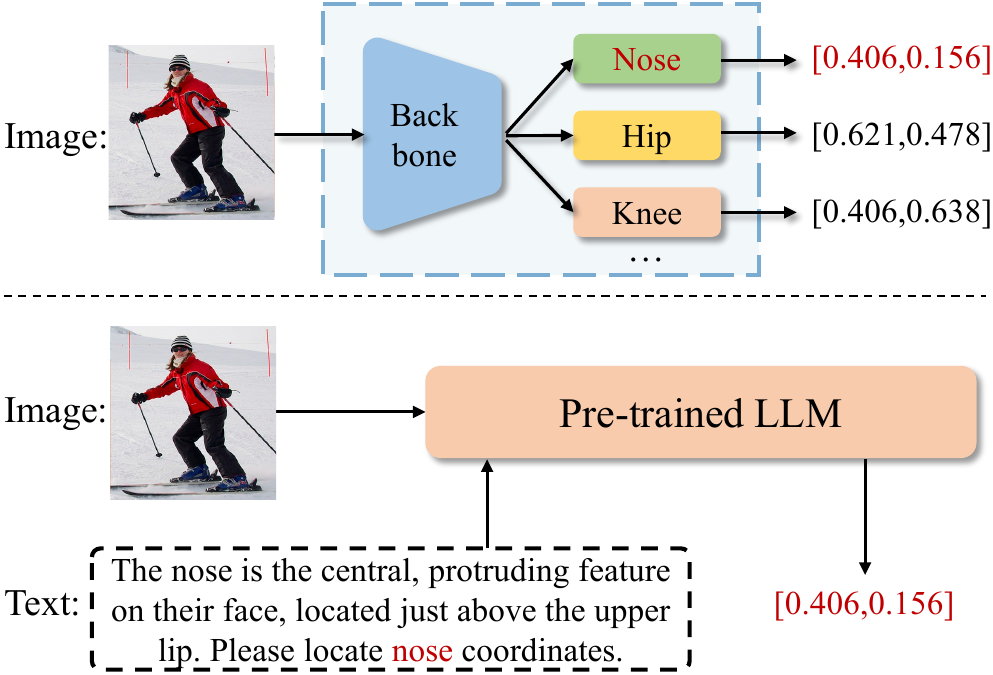}

   \caption{Upper: The conventional keypoint localization methods~\cite{xiao2018simple,xu2022vitpose,li2021human} encodes keypoint prior provided by the training set into model architecture and refers to encoded prior for keypoint localization. Bottom: The proposed LLM-based keypoint localization method refers to keypoint type, location, and relationship descriptions, and utilizes pre-trained powerful LLM~\cite{zhu2023minigpt, liu2023visual} to predict keypoint coordinates. Our method is more general to locate novel keypoints cross datasets, as textual descriptions can be provided flexibly.}
   \label{fig:intro}
   \vspace{-3mm}
\end{figure}

To alleviate the restriction by the training set and pursue a more general model, this work aims to perform keypoint localization from a different perspective, \emph{i.e.}, by referring to clues in textual descriptions that can be flexibly acquired. Inspired by the Large Language Model (LLM)~\cite{zhu2023minigpt, liu2023visual}, we describe keypoint location through natural language and utilize the powerful reasoning capability of LLM for keypoint localization. Previous localization methods need to refer to encoded keypoint priors in network architecture, which is hard to update. Differently, we explicitly send the keypoint name and location descriptions, along with input image to a LLM. Besides visual clues, this new pipeline allows to flexibly input descriptions of novel keypoints by indicating their type, location, and relationship with other keypoints. It also effectively adopts the reasoning capability of pre-trained LLM, therefore improving the generalization ability of keypoint localization.

The above intuition leads to our LocLLM, the first LLM-based localization model for generalizable keypoint localization. As illustrated in Fig.~\ref{fig:intro}, LocLLM formulates the keypoint localization as a question-answer task, taking both image and text description as the input and outputting keypoint coordinates. LocLLM comprises a visual encoder, a projection layer to bridge image and text modalities and a pre-trained LLM. The visual encoder is responsible for learning image representations. The subsequent projector converts image representations into image tokens, which are combined with text tokens as the input of LLM. To effectively train LocLLM, we construct localization-based instruction conversations on existing keypoint localization benchmarks to connect the keypoint description with corresponding coordinates in input image. A parameter-efficient tuning method is proposed to effectively tune the whole model through instruction conversations.

We conduct extensive experiments on different keypoint localization benchmarks including 2D benchmarks like COCO Keypoint~\cite{lin2014microsoft}, MPII~\cite{toshev2014deeppose} and Human-Art~\cite{ju2023human}, and 3D benchmark Human3.6M~\cite{ionescu2013human3}. On standard COCO Keypoint benchmark, LocLLM achieves 77.4 AP, which is comparable with existing SoTA CNN and ViT-based localization methods. LocLLM can also detect 3D human keypoint and achieves promising performance on Human3.6M benchmark. Moreover, LocLLM shows superior generalization ability under various settings. On the cross dataset generalization setup, LocLLM achieves 33.4 PCKh@0.1 on MPII, which is better than ViTPose~\cite{xu2022vitpose} by 7.8. On HumanArt, our method obtains 64.8 AP, outperforming previous methods by a large margin. Moreover, LocLLM can detect novel types of keypoint such as \emph{pelvis} and \emph{neck} through text descriptions, which are not included in the training set. All above experiments demonstrate that LocLLM is a superior generalizable keypoint localization method.

To the best of our knowledge, LocLLM is the first LLM-based keypoint localization model that explicitly exploits natural language description of keypoint into localization. This design allows us to utilize the rich clues from flexible text description and leverage pre-trained LLM for location reasoning. LocLLM achieves promising performance on standard 2D/3D human keypoint localization benchmarks, cross dataset generalization, and novel keypoint detection. LocLLM also incorporates keypoint localization into Multi-modal LLM, which enhances its capability in more fine-grained visual content analysis.

\section{Related Work}
\label{sec:relatedwork}

\subsection{Human Keypoint Localization}
Human keypoint localization aims to locate the person keypoints from input RGB images and plays an important role in computer vision and graphics. Existing keypoint localization methods can be divided into two categories: heatmap-based and regression-based methods.

Heatmap-based keypoint localization encodes keypoint location with a probability map~\cite{tompson2014joint}. This type of methods estimates heatmaps and retrieves keypoint coordinates with a post-processing operation. Currently, heatmap-based methods dominate the field of keypoint localization because heatmap is easy to learn with CNN or Vision Transformer. Pioneer works~\cite{xiao2018simple, sun2019deep,newell2016stacked} design powerful CNN models to estimate high resolution heatmaps for human pose estimation and facial landmark detection. From estimated heatmaps, the target keypoint can be simply obtained by  a post-processing shifting~\cite{newell2016stacked,zhang2020distribution}.

Regression-based keypoint localization directly outputs keypoint coordinates from input image via a neural network, which is adopted by several classical methods~\cite{toshev2014deeppose,carreira2016human}. Many works have been proposed to improve the performance of direct regression. The first kind of methods changes the way of regression. Soft-armgax~\cite{sun2018integral} and Sampling-argmax~\cite{li2021localization} regress keypoint locations by integrating a latent heatmap, which is proved to be superior to direct regression. The second kind of methods improves regression by proposing new loss functions. RLE~\cite{li2021human} changes the predefined Gaussian or Laplace distribution in commonly used regression loss with a learned distribution via normalizing flow. Finally, researchers also propose more powerful backbones to improve the performance of direct regression, such as TokenPose~\cite{li2021tokenpose} and PETR~\cite{shi2022end}.

All above methods directly encode keypoint type clues into architecture and implicitly learn the keypoint location through training data. Therefore, their generalization capability is restricted by the model architecture and training data. In contrast, our method explicitly exploits keypoint type and location from language description and LLM, making it more generalizable to detect novel keypoints.

\subsection{Multi-modal Large Language Model}
Large Language Model (LLM) shows remarkable reasoning capabilities in natural language processing tasks, therefore researchers try to enhance it with additional modalities, \emph{e.g.}, image, audio, motion, \emph{etc.}, to develop Multi-modal LLM (MLLM). Flamingo~\cite{alayrac2022flamingo} proposes Perceiver to extract representative visual tokens and add them into LLM through cross-attention. BLIP-2~\cite{li2023blip} proposes Q-Former to align visual features with text tokens in LLM. Instruction tuning~\cite{wei2021finetuned} is a commonly adopted way to align vision and language modalities to improve the ability of MLLM. Mini-GPT4~\cite{zhu2023minigpt} and LLaVA~\cite{liu2023visual} construct a high-quality
instruction tuning dataset and fine-tune only a single fully connection layer to construct MLLMs. Instruct-BLIP~\cite{dai2023instructblip} introduces an instruction-aware visual feature extraction method and fine-tunes the entire Q-Former, showing promising zero-shot performance on various multi-modal tasks. mPlug-Owl~\cite{ye2023mplug} incorporates a visual abstractor to align the two-modalities, and fine-tune both the visual encoder and visual abstractor during the pre-training stage. AnyMAL~\cite{moon2023anymal} aligns not only image but also more modalities, such as video, audio and IMU motion sensor to LLM.

\subsection{Large Language Model for Vision Tasks}
Despite remarkable progress in MLLM, most methods still focus on vision-language tasks, such as VQA and image caption. The effectiveness of LLM in classical vision tasks, \emph{e.g.}, detection, segmentation and localization has not been fully exploited. LISA~\cite{lai2023lisa} defines a new vision task named as reasoning segmentation and propose a framework to extract referring text embedding from MLLM, which is sent to a segmentation model like SAM~\cite{kirillov2023segment} to perform segmentation. VisionLLM~\cite{wang2023visionllm} proposes a framework to address vision tasks such as detection and segmentation through LLM with a complex image tokenizer. However, none of above work exploits the effectiveness of LLM in keypoint localization task, which requires locating target in sub-pixel level accuracy, rather than the coarse object level in detection and segmentation. To the best of our knowledge, this is the first work that exploits LLM for keypoint localization and demonstrates that LLM can achieve superior performance in generalizable keypoint localization.

\section{Method}
\label{sec:method}

\begin{figure*}[t]
  \centering
   \includegraphics[width=0.9\linewidth]{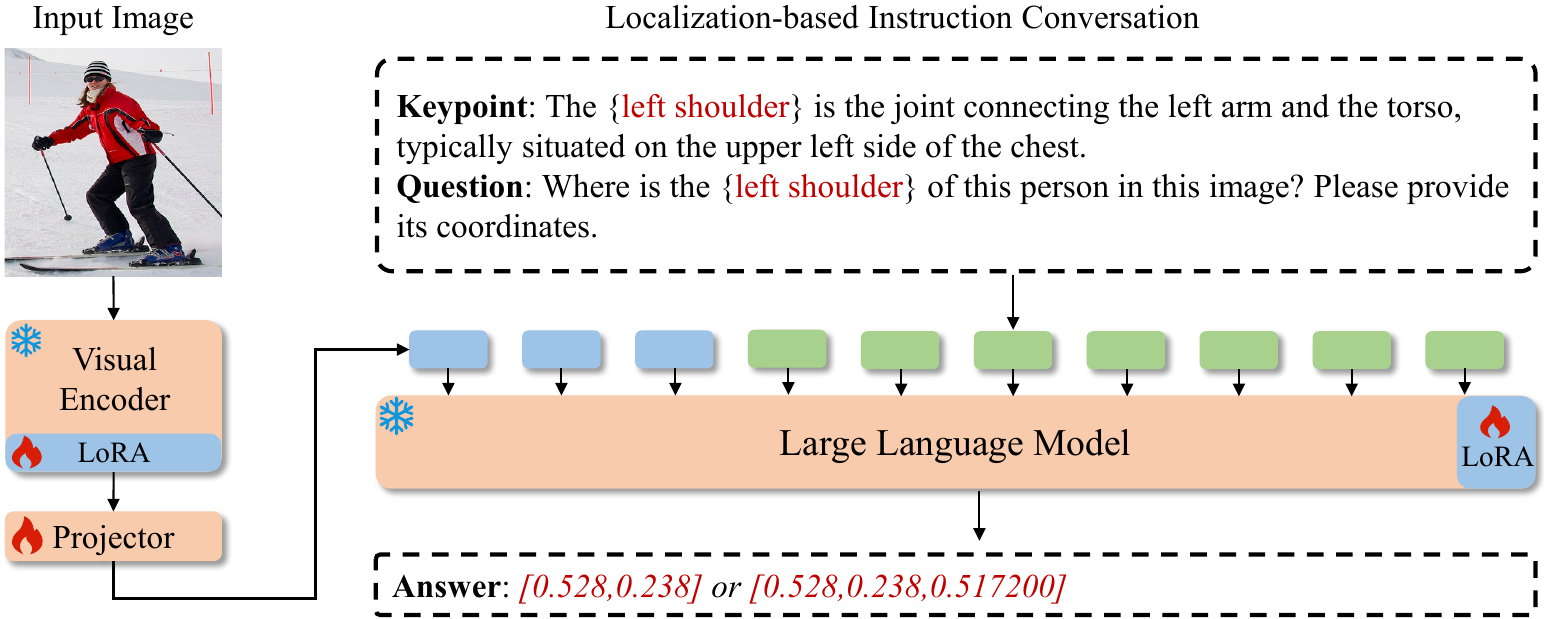}

   \caption{The proposed LocLLM for keypoint localization via large language model. LocLLM takes image and text instruction as input and contains three parts: a visual encoder, a projector and a decoder-only LLM. The image input is processed by visual encoder and projector to extract image tokens. The LLM takes the image tokens and text tokens as input and output corresponding keypoint coordinates. During training, we freeze the visual encoder and LLM and only update a small set of learnable parameters with projector, therefore relieving the training cost.}
   \label{fig:framework}
   \vspace{-2mm}
\end{figure*}

\subsection{Overview}
The goal of human keypoint localization is to estimate the coordinates of target keypoints from input image, which can be conceptually denoted as,
\begin{equation}\label{eq:form1}
  \{\mathcal K_i\}_{i=1}^n = \operatorname{locate}(\mathcal I),
\end{equation}
where $\mathcal K_i$ denotes the coordinates of the $i$-th type keypoint, \emph{e.g.}, person shoulder or knee, $n$ is the total type of keypoint defined in each dataset. Following Eq.~\ref{eq:form1}, previous methods~\cite{xiao2018simple,li2021human} encode keypoint clues into network architecture and learn location prior from training set, which limits their generalization ability.

Different from above formulation, in this paper we investigate keypoint location description and utilize the powerful large language model to perform localization, rewriting Eq.~\eqref{eq:form1} into
\begin{equation}\label{eq:form2}
  \{\mathcal K_i\} = \operatorname{locate}(\mathcal I, \mathcal T_i),
\end{equation}
where $\mathcal T_i$ contains the text description of $i$-th target keypoint. In this way, the keypoint type is not solely encoded into the model but also through the text description input, allowing us to explicitly exploit keypoint type, location and relationship, and even detect novel keypoint.

Following Eq.~\eqref{eq:form2} and most MLLM work~\cite{liu2023visual,xuan2023pink}, we formulate the keypoint localization as a visual question answer (VQA) task and utilize the powerful reasoning capability of LLM to achieve our goal. Specifically, we propose LocLLM, a generative model that aims to complete multi-modal sentences to output keypoint coordinates. As shown in Fig.~\ref{fig:framework}, LocLLM consists three main components: a visual encoder $\Phi_{V}(\cdot)$, a linear projector $\Phi_{P}(\cdot)$ and a large language model $\Phi_{L}(\cdot)$. The input of LocLLM contains two parts, the image $\mathcal I \in \mathbb R^{3\times H\times W}$ and text instruction $\mathcal T$. The output is target keypoint coordinates $\mathcal K$, this process can be denoted as,

\begin{equation}\label{eq:locllm}
  \mathcal K = \Phi_{L}(\Phi_P(\Phi_V(\mathcal I)), \mathcal T).
\end{equation}

The visual encoder $\Phi_{V}(\cdot)$ takes an image $\mathcal I \in \mathbb R^{3\times H\times W}$ as input and outputs a sequence of image features $\mathcal F = (f_1, f_2, ..., f_m)$, where $m$ is the number of image features. The image features are further projected into image tokens by a single linear layer projector $\Phi_{P}(\cdot)$~\cite{liu2023visual}, \emph{i.e.},
\begin{equation}
 \{v_1, v_2, ..., v_m\}=\Phi_P(\Phi_V(\mathcal I)).
\end{equation}

The text $\mathcal T$ will also be processed by the tokenizer of $\Phi_L(\cdot)$ to generate text tokens $\{t_1, ..., t_l\}$, which is combined with image tokens to be sent to $\Phi_{L}(\cdot)$. Utilizing the self-attention mechanism, the LLM is capable of understanding the contextual relationships between different types of tokens, enabling it to generate responses based on both text and image inputs. Formally, the output of LLM $\Phi_{L}(\cdot)$ is also a sequence, \emph{i.e.},
\begin{equation}
 \{k_1, k_2, ..., k_s\}=\Phi_L(\{v_1,..., v_m, t_1, ..., t_l\}),
\end{equation}
where $s$ denotes the length of output tokens, $k_i$ is generated sequentially based on all previous tokens $\{v_1,..., v_m, t_1, ..., t_l, k_1, ..., k_{i-1}\}$. Then $k_i$ will be mapped to LLM vocabulary by a linear classifer $\mathcal C(\cdot)$.  During training, we encode the keypoint coordinates $\mathcal K$ into ground truth vocabulary class sequence $\{k_1^*, k_2^*, ..., k_s^*\}$ and add standard Cross Entropy loss on the output classification score, which can be denoted as,
\begin{equation}
  \mathcal L = \sum_i \textbf{CE}(\mathcal C(k_i), k_i^*).
\end{equation}
During inference, we decode the output tokens to vocabulary words by selecting the words with the highest probability in $\mathcal C(k_i)$ to get estimated keypoint coordinates.

Previous works~\cite{liu2023visual,xuan2023pink,zhu2023minigpt} reveal that the text instruction $\mathcal T$ plays a key role in unleashing the power of LLM to complete corresponding tasks. Moreover, how to effectively tune LocLLM to generate accurate keypoint coordinates is also a challenge. Therefore, following two parts proceed to introduce the detailed localization-based instruction conversation construction to instruct LLM to perform keypoint localization, and a parameter-efficient tuning pipeline to train the LocLLM, respectively.

\subsection{Localization-based Instruction Conversation}
Constructing proper instructions is a key step to tune LLM towards a specific task, which is verified in many visual instruction tuning methods~\cite{liu2023visual,zhu2023minigpt}. To enable LLM perform keypoint localization accurately, we create the following localization-based instruction conversation as LocLLM input. The instruction template is shown in Table~\ref{tab:conversation} and an example can be found in Fig.~\ref{fig:framework}.

\begin{table}[t]
\setlength{\tabcolsep}{5.5mm}{
\small
\begin{center}
\begin{tabular}{l|c}
\toprule
\multicolumn{2}{l}{\textbf{Instruction Template}} \\
\midrule
    \textbf{Image}: & \{image tokens\} \\
    \textbf{Keypoint}: & \{keypoint location description\} \\
    \textbf{Question}: & \{question to perform localization\}  \\
    \textbf{Answer}: & \{keypoint coordinates\}\\
    \bottomrule
\end{tabular}
\end{center}
}
\vspace{-2mm}
\caption{Illustration of the proposed localization-based instruction conversation template.}
\vspace{-2mm}
\label{tab:conversation}
\end{table}

\textbf{Keypoint Description.} Different from previous work that only provide question, we additionally provide a sentence that describes the keypoint location on human body to help LLM locate target keypoint. For each type of keypoint, we ask ChatGPT to generate corresponding description and manually check it with Wikipedia. Manual intervention aims to ensure the descriptions are reliable. The generation and manual intervention are offline and no longer needed once the descriptions are checked. The detailed description of each keypoint can be found in supplemental materials and its effectiveness is verified in Sec.~\ref{sec:experiments}.

\textbf{Keypoint Coordinates Format.} One challenge in localization-based instruction conversation is how to format the keypoint coordinates so that LLM can predict them accurately. According to previous work OFA~\cite{wang2022ofa}, Shikra~\cite{chen2023shikra} and Pink~\cite{xuan2023pink}, we investigate two types of keypoint coordinates format in instruction conversation.

The first is adopting location token to represent keypoint coordinates, which is used in many methods such as OFA~\cite{wang2022ofa}. For example, a keypoint with $(95, 123)$ coordinates can be converted into two $\langle 095 \rangle$ $\langle 123 \rangle$ tokens. Considering that the image size is fixed, \emph{e.g.}, $224\times 224$, we can add a set of fixed location token into tokenizer to represent keypoint coordinate. However, the drawback of location token is that their embedding should be additionally learned, and it is hard to represent decimal coordinate such as depth in 3D keypoint representation.

The second is to directly adopt decimal string to represent keypoint coordinates. Specifically, we normalize the keypoint coordinates into the range $[0,1]$ with respect to the image size in spatial dimension or camera 3D bounding box size in depth dimension and preserve 3/6 decimal places for each number, \emph{i.e.},
\begin{equation}\label{eq:kptstring}
  [0.abc,0.def,0.ghijkl],
\end{equation}
where lowercase letters denote any number between $0$ and $9$. For decimal string the tokenizer will split it into a set of words, \emph{e.g.}, $"0.abc"$ into $\{"0", ".", "a", "b", "c"\}$, which is already included in LLM vocabulary. Therefore, the advantage of the second format is that we do not need to add new tokens into LLM vocabulary and train corresponding embedding layer. Moreover, the decimal string format allow us to easily extend the framework to locate 3D keypoint with minimal modification, \emph{i.e.}, just extend the string to include depth dimension.

\textbf{Multi-Round Conversation.} One image may contain multiple target keypoints, \emph{e.g.}, knee and shoulder. Therefore it is not efficient to ask only one keypoint location in one conversation. To boost the training efficiency, we follow the VQA method to construct multi-round conversation to ask multiple keypoint locations for one input image in a single forward pass.

\subsection{Parameter-Efficient Tuning}
Due to the huge parameters of LLM, it is not feasible to fine-tune the entire model with limited GPU resource. Moreover, fully fine-tuning LLM and visual encoder requires millions of image-text pairs to avoid model collapse, which is unrealistic in keypoint localization task.

To perform an efficient training and enable the entire model to benefit from multi-modal localization-based instruction conversation, we freeze the visual encoder and LLM and introduce a small set of learnable parameters into them. This approach prevents the visual encoder and LLM from suffering semantic loss due to the limited instruction text data and provide a parameter-efficient training way to perform keypoint localization. Specifically, we adopt LoRA~\cite{hu2021lora} to training LocLLM. Given a weight matrix $W\in \mathbb R^{d\times k}$ in pre-trained model , the LoRA is defined as follows,
\begin{equation}
  \hat W = W + \Delta W = W + W_{B}W_{A},
\end{equation}
where $W_{A}\in \mathbb R^{r\times k}$ and $W_{B}\in \mathbb R^{d\times r}$ denote the weight matrices of LoRA module, $r$ is the hidden dimension which is much smaller than $d$ and $k$. $W_{B}$ is initialized to zero to ensure that at the beginning of the training LoRA does not change the original output.

We add LoRA modules into both visual encoder and LLM, and fine-tune them with the linear projector. In our method, the whole trainable parameters are 8.7M, which is much smaller than the whole model and common CNN and ViT models. Following previous methods~\cite{liu2023visual}, LocLLM is trained in two stages. In this first stage, we align the image and text by only fine-tuning the projection layer on image-text pairs CC3M~\cite{sharma2018conceptual}. In the second stage, we freeze the visual encoder and LLM and fine-tune the newly added LoRA module and projector on the localization-based instruction conversations. Therefore, LocLLM can benefit from the multi-modal conversation and perform keypoint localization accurately.

\subsection{Baseline: CLIP-based Keypoint Localization}
\begin{figure}[t]
  \centering
   \includegraphics[width=1.0\linewidth]{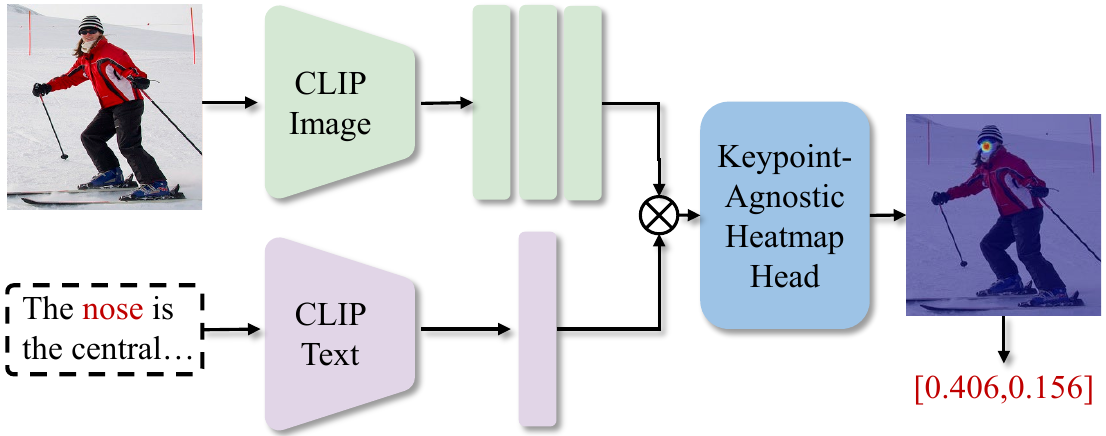}

   \caption{Illustration of the CLIP-based keypoint localization. }
   \label{fig:clip}
   \vspace{-2mm}
\end{figure}

Another potential way to utilize keypoint description to guide localization is to adopt vision-language model such as CLIP~\cite{radford2021learning}. Different from LLM, CLIP aligns the image and text feature space through millions of image-text pairs, so that we can extract text feature and use it to guide image feature extraction in conventional localization method. Therefore, we also propose a simple CLIP-based keypoint localization baseline to compare with LocLLM, with the aim to indicate that LLM is important in utilizing keypoint description for localization.

As shown in Fig.~\ref{fig:clip}, we adopt CLIP image and text encoders to extract corresponding features from input image and text, which can be denoted as $\mathcal F_v$ and $f_t$. A text-conditioned feature map can be obtained by element-wise multiplication above two features, \emph{i.e.}, $\mathcal F_v^t=\mathcal F_v \odot f_t$, which is sent to a keypoint-agnostic head to estimate corresponding heatmap. Note that the proposed baseline is different from CLAMP~\cite{zhang2023clamp}, which also adopts CLIP to locate keypoints. CLAMP only uses text to enhance the feature, and still uses $n$-channel heatmap head to estimate $n$ heatmaps defined by training data. Therefore, it cannot be used to locate novel type keypoints that are out of training set. In contrast, the proposed CLIP baseline introduces the text-conditioned feature and keypoint-agnostic head to locate keypoint, thus is not limited to the fixed keypoint set in training set. Performance of this baseline is tested in Sec.~\ref{sec:experiments}. 

\section{Experiments}
\label{sec:experiments}

\subsection{Datasets and Evaluation Metrics}

We conduct experiments on different datasets, including image-text pair dataset CC3M~\cite{sharma2018conceptual}, 2D human keypoint localization datasets COCO Keypoint~\cite{lin2014microsoft}, MPII~\cite{toshev2014deeppose} and HumanArt~\cite{ju2023human}, and 3D human keypoint localization dataset Human3.6M~\cite{ionescu2013human3}.

Filtered CC3M~\cite{sharma2018conceptual} is constructed by LLaVA~\cite{sharma2018conceptual} and is the widely adopted visual instruction tuning dataset. It contains 595K image-text pairs. We adopt this dataset for the first stage training of LocLLM. 

COCO Keypoint~\cite{lin2014microsoft} contains 64K images of 270K persons labeled with 17 keypoints. Its \texttt{train} set contains 57K images, 150K persons. The \texttt{val} set contains 5K images, 6.3K persons is used for evaluation. We adopt this dataset to construct 2D localization-based instruction conversation to train LocLLM at the second stage.

Human3.6M~\cite{ionescu2013human3} is a large scale indoor benchmark for 3D human keypoint localization, which consists of 3.6 million  images  from 4 camera views. Following the standard protocols, We adopt subjects 1, 5, 6, 7, 8 to construct 3D localization-based instruction conversation to training LocLLM at the second stage and test model on subjects 9, 11. Besides above datasets, we further evaluate the generalization ability of LocLLM on other human keypoint localization datasets, \emph{e.g.}, MPII~\cite{toshev2014deeppose} and HumanArt~\cite{ju2023human}.

We follow the standard evaluation metric to report performance on each dataset. We report PCKh@0.5/0.1 on MPII and mAP on other 2D datasets. For 3D human keypoint localization, we report the Mean Per Joint Position Error (MPJPE) to evaluate the error of each method.

\subsection{Implementation Details}

\textbf{Model Architecture.}
We adopt ViT-L/14 as visual encoder, which is pre-trained with DINOv2~\cite{oquab2023dinov2} weights. For pre-trained LLM, we adopt an instruction-tuned model vicuna-7B~\cite{vicuna2023}.
The projection layer is a single fully connection layer. The LoRA modules are inserted into the q and v of each self-attention layer of both visual encoder and LLM, with a hidden dimension $r=8$.

\textbf{Training Details.}
AdamW is adopted as the optimizer. In the first stage, the model is trained for 1 epoch with a batch size of 128 and weight decay of 0.0. After a warm-up period of 200 steps,
the learning rate starts at 0.03 and decays to 0 with the cosine schedule. In the second stage, the
model is trained on COCO Keypoint for 3 epochs for ablation study and 12 epochs for final comparison on 2D human keypoint localization. For 3D keypoint localization, we follow RLE~\cite{li2021human} to train model on mixed Human3.6M and MPII. The model is trained with a batch size of 64 by gradient accumulation and weight decay of 0.05. The warm-up phase consists of 10k steps and the learning rate starts at 5e-4. The input image is resized to 224$\times$224. Note
that our model has only 8.7M trainable parameters, making it feasible to train with consumer GPUs,
e.g., four 24G NVIDIA 3090s.

\subsection{Ablation Study}

\textbf{Localization-based Instruction Conversation.} We first analyze each component in constructing localization-based instruction conversation. Results are shown in Table~\ref{tab:kptformat}. The keypoint description is useful to help LocLLM to locate keypoints. As discussed in Sec.3.2, there are two ways to represent keypoint coordinates, \emph{i.e.}, discrete location token and decimal string. As shown in Table~\ref{tab:kptformat}, we observe that decimal string is superior to location token. Observing the loss we can conclude that introducing location token into LLM requires to retrain the embedding layer, which is hard to learn from a small scale dataset.

We also investigate the effectiveness of conversation round. Training LocLLM with single-round conversation achieves 68.9 AP on COCO \texttt{val} set, this can be improved to 72.4 AP when adopting a multi-round conversation paradigm during training. This indicates that providing more examples can help the model to perform better on human keypoint localization task.

\begin{table}[t]
\setlength{\tabcolsep}{2.3mm}{
\small
\centering
\begin{tabular}{l|ccccc}
\toprule
Component & $\operatorname{AP}$ & $\operatorname{AP}^{50}$ & $\operatorname{AP}^{75}$ & $\operatorname{AP}^{M}$ & $\operatorname{AP}^L$\\
\midrule
\multicolumn{6}{l}{\emph{Keypoint Description}} \\
\midrule
No description & 70.8 & 91.4 & 78.0 & 68.0 & 75.8\\
With description & \textbf{72.4} & \textbf{92.4} & \textbf{80.1} & \textbf{69.2} & \textbf{77.4}\\
\midrule
\multicolumn{6}{l}{\emph{Keypoint Format}} \\
\midrule
Location token & 67.2 & 91.4 & 75.9 & 64.2 & 71.9\\
Decimal string & \textbf{72.4} & \textbf{92.4} & \textbf{80.1} & \textbf{69.2} & \textbf{77.4}\\
\midrule
\multicolumn{6}{l}{\emph{Conversation Round}} \\
\midrule
Single & 68.9 & 92.4 & 76.6 & 66.2 & 73.1\\
Multiple & \textbf{72.4} & \textbf{92.4} & \textbf{80.1} & \textbf{69.2} & \textbf{77.4}\\
\bottomrule
\end{tabular}
}
\caption{Component analysis of localization-based instruction conversation on COCO \texttt{val} set.}
\label{tab:kptformat}
\end{table}

\begin{table}[t]
\setlength{\tabcolsep}{1.6mm}{
\small
\centering
\begin{tabular}{ccc|ccccc}
\toprule
$\Phi_V(\cdot)$ & $\Phi_P(\cdot)$ & $\Phi_L(\cdot)$ & $\operatorname{AP}$ & $\operatorname{AP}^{50}$ & $\operatorname{AP}^{75}$ & $\operatorname{AP}^{M}$ & $\operatorname{AP}^L$\\
\midrule
 & \checkmark &  & 39.5 & 77.1 & 36.2 & 38.5 & 40.9\\
 \checkmark & \checkmark &  & 70.3 & 92.3 & 78.0 & 67.6 & 74.4\\
 & \checkmark & \checkmark & 55.1 & 87.0 & 60.2 & 52.9 & 58.3\\
 \checkmark & \checkmark & \checkmark & \textbf{72.4} & \textbf{92.4} & \textbf{80.1} & \textbf{69.2} & \textbf{77.4}\\
\midrule
\multicolumn{3}{l|}{Only second stage} & 70.6 & 92.0 & 78.9 & 67.8 & 75.3\\
\bottomrule
\end{tabular}
}
\caption{Ablation study on parameter-efficient tuning each compoint of LocLLM on COCO \texttt{val} set.}
\vspace{-2mm}
\label{tab:ab-component}
\end{table}

\textbf{Paramter-Efficient Tuning.} The way of tuning the LLM is also important to the final performance. Due to the limited GPU resource and annotation, we could not fine-tune the whole model. Therefore, we insert some leanable parameter modules into the model to conduct parameter-efficient tuning. This experiment investigates the effects of insert locations of learnable parameter module to the performance of LocLLM.  As shown in Table~\ref{tab:ab-component}, only training projector layer cannot learn much information from the data and performs badly on COCO \texttt{val} set. Inserting learnable module into either visual encoder or LLM can both substantially improve the localization performance. Among them, we find that tuning both visual encoder and LLM achieves the best performance.

\subsection{2D/3D Human Keypoint Localization}

\begin{table}[t]
  \centering
  \small
  \setlength{\tabcolsep}{1.7mm}{
\begin{tabular}{lcccccc}\toprule
\multicolumn{1}{l}{{Method}} & \multicolumn{1}{c}{{AP}} & \multicolumn{1}{c}{{AP$^{50}$}} & \multicolumn{1}{c}{{AP$^{75}$}} & \multicolumn{1}{c}{{AP$^{M}$}} & \multicolumn{1}{c}{{AP$^{L}$}} & \multicolumn{1}{c}{{AR}} \\ \midrule
\multicolumn{7}{l}{\emph{Heatmap-based}} \\ \midrule
\multicolumn{1}{l}{SimplePose~\cite{xiao2018simple}}&  74.4 & 92.6 & 82.5 & 71.5 & 79.2 & 77.6  \\
\multicolumn{1}{l}{HRNet~\cite{sun2019deep}} &  76.8 & 93.6 & 83.6 & 74.0 & 81.5 & 79.6\\
SimCC~\cite{li2022simcc} & 76.5 & 93.2 & 83.1 & 73.6 & 81.5 & 79.7\\
ViTPose~\cite{xu2022vitpose} & 77.4 & 93.6 & 84.8 & \textbf{74.7} & \textbf{81.9} & 80.2\\
\midrule

\multicolumn{7}{l}{\emph{Regression-based}} \\ \midrule
\multicolumn{1}{l}{DeepPose~\cite{toshev2014deeppose}}& 53.8 & 82.6 & 59.2 & 52.2 & 57.3 & 66.8 \\
\multicolumn{1}{l}{RLE~\cite{li2021human}} & 74.0 & 91.5 & 81.6 & 70.9 & 78.5 & 76.8 \\
 \midrule
\multicolumn{7}{l}{\emph{Language-based}} \\ \midrule
CLIP baseline &  73.1 & 92.5 & 81.3 & 70.3 & 77.4 & 76.5\\
Ours (LocLLM) & \textbf{77.4} & \textbf{94.4} & \textbf{85.2} & {74.5} & {81.8} & \textbf{80.6}\\
\bottomrule
\end{tabular}
}
\caption{Comparison with other methods on COCO Keypoint \texttt{val} set in 2D keypoint localization. All results are obtained by evaluating the official model weights provided by the authors using GT bbox without flip test.}
\label{tab:coco2d}
\end{table}

\begin{table}[t]
\centering
\small
\setlength{\tabcolsep}{2.1mm}{
\begin{tabular}{lcccccc}\toprule
\multicolumn{1}{l}{{Method}} &  Eat & Pose & Sit & Wait & Walk & Avg \\ \midrule
Sun \emph{et al.}~\cite{sun2017compositional} & 54.2 & 53.1 & 71.7 & 53.4 & 47.1 & 59.1 \\
PoseNet~\cite{moon2019camera} & 50.1 & 46.8 & 61.9 & 49.9 & 41.8 & 53.3 \\
Sun \emph{et al.}~\cite{sun2018integral} & 49.5 & 43.8 & 58.9 & 47.8 & 38.9 & 49.6 \\
RLE~\cite{li2021human} &  44.5 & 43.1 & 59.2 & 44.1 & \textbf{37.5} & 48.6\\
\midrule
Ours (LocLLM) & \textbf{41.2} & \textbf{40.0} & \textbf{53.6} & \textbf{41.8} & {37.8} & \textbf{46.6} \\
\bottomrule
\end{tabular}
}
\caption{Comparison with other methods on Human3.6M benchmark in monocular 3D keypoint localization.}
\label{tab:h36m3d}
\vspace{-2mm}
\end{table}

\begin{table*}
\centering
\setlength{\tabcolsep}{2.8mm}{
\small
\begin{tabular}{l|cccccc|cccccc}
  \toprule
  \multirow{2}{*}{Method} & \multicolumn{6}{c|}{Human-Art}  & \multicolumn{6}{c}{MPII} \\
  \cmidrule{2-13}
  & AP & AP$^{50}$ & AP$^{75}$ & AP$^{M}$ & AP$^{L}$ & AR & Shou. & Elbo. & Hip & Knee & Mean & Mean0.1\\
  \midrule
  SimplePose~\cite{xiao2018simple} & 48.4 & 73.0 & 50.7 & 27.2 & 50.7 & 52.8 & 93.7 & 85.6 & 85.4 & 81.6 & 84.7 & 22.2\\
  SimCC~\cite{li2022simcc} & 51.7 & 75.2 & 54.8 & 26.2 & 54.3 & 57.0 & 92.2 & 84.1 & 82.8 & 80.5 & 83.2 & 28.5 \\
  HRNet~\cite{sun2019deep} & 53.4 & 76.3 & 56.5 & 30.4 & 55.9 & 57.5 & 93.4 & 86.1 & 85.0 & 81.9 & 85.8 & 26.9\\
  ViTPose~\cite{xu2022vitpose} & 53.8 & 77.9 & 57.4 & 31.4 & 56.6 & 58.7 & 94.5 & 88.2 & 87.3 & 85.0 & 86.9 & 25.8\\
  \midrule
  CLIP baseline & 49.3 & 75.7 & 51.8 & 27.7 & 52.0 & 54.5 & 95.5 & 88.8 & 87.5 & 84.7 & 87.1 & 25.0\\
  Ours & \textbf{64.8} & \textbf{87.4} & \textbf{70.4} & \textbf{40.9} & \textbf{67.4} & \textbf{69.3} & \textbf{96.1} & \textbf{90.3} & \textbf{89.8} & \textbf{88.0} & \textbf{89.3} & \textbf{33.4}\\
  \bottomrule
\end{tabular}
}
\caption{Comparison with other methods on cross dataset generalization on Human-Art and MPII. All models are trained on COCO Keypoint \texttt{train} set. We only evaluate the accuracy of keypoints that appear in COCO Keypoint.}
\label{tab:crossdataset}
\end{table*}

This section demonstrates that LLM can perform well on conventional keypoint localization tasks, including 2D human pose estimation and 3D human pose estimation. We compare LocLLM with recent methods on COCO Keypoint for 2D keypoint localization and Human3.6M for 3D keypoint localization. Results are shown in Table~\ref{tab:coco2d} and Table~\ref{tab:h36m3d}.

We first compare LocLLM with other methods in 2D human pose estimation task and report performance on COCO Keypoint \texttt{val} set. As shown in Table~\ref{tab:coco2d}, existing 2D human pose estimation methods can be divided into heatmap-based method: SimplePose~\cite{xiao2018simple}, HRNet~\cite{sun2019deep}, SimCC~\cite{li2022simcc} and ViTPose~\cite{xu2022vitpose}, and regression-based methods, including: DeepPose~\cite{toshev2014deeppose} and RLE~\cite{li2021human}. Our method can be viewed as a language-based method that utilize text keypoint description to locate keypoint position, therefore we also report the performance of CLIP baseline in Fig.~\ref{fig:clip}. As shown in Table~\ref{tab:coco2d}, our method achieves superior performance on COCO \texttt{val} set, which is comparable to recent SoTA methods such as ViTPose and RLE. With a few learnable parameters (8.7 M), LocLLM can achieve comparable performance with recent SoTA methods.

In Table~\ref{tab:h36m3d} we further demonstrate that LocLLM can perform 3D keypoint localization in monocular RGB image, which is rarely exploited by previous MLLM methods. Benefited by the decimal string representation of keypoint coordinates, LocLLM can be easily extended to 3D keypoint localization by simply adding a depth dimension in outputs. We follow RLE~\cite{li2021human} to conduct experiments and compare with previous methods. As shown in Table~\ref{tab:h36m3d}, LocLLM achieves superior 3D keypoint localization performance, indicating its capacity in depth understanding.

\subsection{Cross Dataset Generalization}
This section aims to evaluate the generalization ability of LocLLM on locating keypoint for unseen human pose from other datasets. We conduct cross dataset validation to test LocLLM trained on COCO on various different human pose estimation benchmarks such as Human-Art~\cite{ju2023human} and MPII~\cite{toshev2014deeppose}. The results are shown in Table~\ref{tab:crossdataset}.

Human-Art is a benchmark that contains human pose in both natural scenes such as sports or outdoor, and artificial scenes including cartoon, digital art, ink painting and \emph{etc}. It is suitable to evaluate the generalization ability of keypoint localization methods. As shown in Table~\ref{tab:crossdataset}, we compare LocLLM with previous conventional keypoint localization methods in Table~\ref{tab:coco2d}. Conventional localization methods achieves promising performance on COCO, but suffer a large performance drop on Human-Art. For example, ViTPose achieves 77.4 AP on COCO, but only obtains 53.8 AP on Human-Art. In contrast, our method achieves 64.8 AP on Human-Art, significantly better than compared methods. This indicates the superior generalization ability of our method in cross dataset validation.

\subsection{Novel Keypoint Localization}

\begin{table}[t]
\setlength{\tabcolsep}{2.4mm}{
\small
\centering
\begin{tabular}{l|ccccc}
\toprule
Method & $\operatorname{AP}$ & $\operatorname{AP}^{50}$ & $\operatorname{AP}^{75}$ & $\operatorname{AP}^{M}$ & $\operatorname{AP}^L$\\
\midrule
Full keypoint & 72.4 & 92.4 & 80.1 & 69.2 & 77.4 \\
\midrule
CLIP baseline & 43.0 & 89.2 & 28.7 & 42.0 & 44.9\\
Ours & 67.1 & 91.2 & 76.3 & 64.9 & 71.1\\
\bottomrule
\end{tabular}
}
\caption{Results of removing 4 keypoints from training and testing on COCO Keypoint \texttt{val} set.}
\label{tab:novelkeypoint}
\end{table}

\begin{table}
\centering
\setlength{\tabcolsep}{1.8mm}{
\small
\begin{tabular}{l|cccc|cc}
  \toprule
  \multirow{2}{*}{Method} & \multicolumn{4}{c|}{Seen}  & \multicolumn{2}{c}{Unseen} \\
  \cmidrule{2-7}
  & Shou. & Elbo. & Hip & Knee & Pelvis & Neck\\
  \midrule
  CLIP baseline & 95.5 & 88.8 & 87.5 & 84.7 & 1.7 & 5.6\\
  Ours & 96.1 & 90.3 & 89.8 & 88.0 & \textbf{43.6} & \textbf{36.9}\\
  \bottomrule
\end{tabular}
}
\caption{Comparison with other methods on novel keypoint localization on MPII.}
\label{tab:novelkptmpii}
\vspace{-2mm}
\end{table}

We finally show that LocLLM can even locate novel type of keypoint that are never seen during training, to demonstrate its superior generalization ability. Existing methods encode the keypoint prior into the network architecture, making them hard to generalize to unseen keypoints. Our LocLLM is not subject to this restriction and can locate novel type of keypoints by referring to text descriptions.

To verify this claim, we conduct two experiments on COCO Keypoint and MPII. For the first experiment, we remove 4 keypoints (\emph{right elbow}, \emph{left wrist}, \emph{left knee}, \emph{right ankle}) from total keypoints (17) during training and do not apply any data-augmentation. In other words, 13 types of keypoints are used for training, and the model is tested to detect 17 keypoints. The results are shown in Table~\ref{tab:novelkeypoint}, where ``Full keypoint'' uses all keypoints for training, hence is regarded as the upper bound. By removing 4 types of keypoint from training, the performance of CLIP basline deteriorates greatly, \emph{e.g.}, from 73.1 to 43.1. In contrast, our LocLLM can still achieve a reasonably good performance.

\begin{figure}[t]
  \centering
   \includegraphics[width=1.0\linewidth]{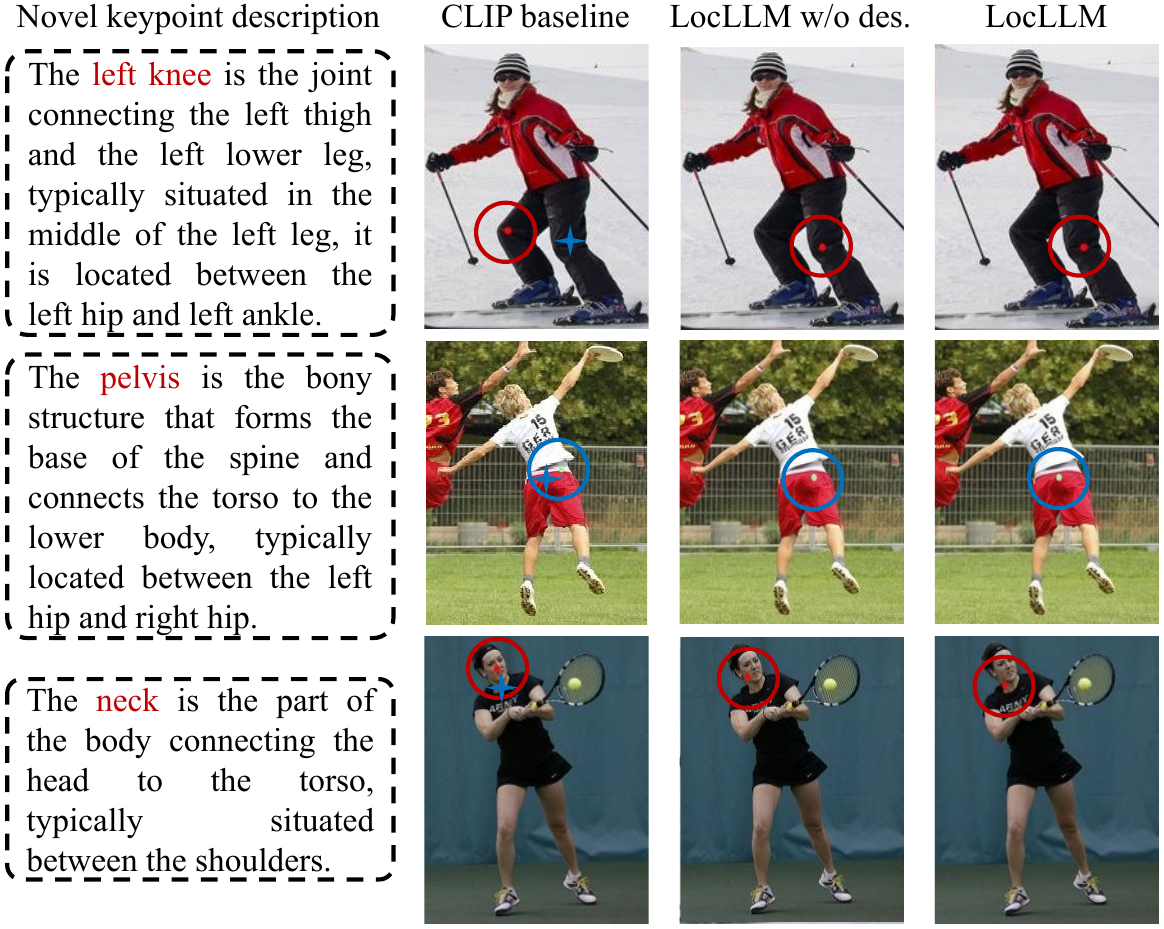}

   \caption{Localization results of three novel keypoints which are not seen during training (denoted by blue star in the first column image). It can be observed that CLIP baseline matches each novel keypoint to a similar keypoint in the training set, \emph{e.g.}, it locates the left knee to right knee, pelvis to right hip, and neck to nose. In contrast, our LocLLM can locate novel keypoint accurately.}
   \label{fig:novelkpt}
   \vspace{-3mm}
\end{figure}

For the second experiment, we use all 17 keypoints of COCO Keypoint for training, then test the model on another dataset with different keypoint definition, \emph{i.e.}, the MPII dataset. Note that, the \emph{Pelvis} and \emph{Neck} keypoints in MPII are not seen by model trained on COCO Keypoint. We hence also report the performance on them. As shown in Table~\ref{tab:novelkptmpii}, our LocLLM achieves 43.6 accuracy on \emph{Pelvis}, which is much better than the CLIP baseline with only 1.7 accuracy. In Fig.~\ref{fig:novelkpt} we show some examples on novel keypoint localization. It can be observed that our method can accurately locate novel keypoints with the help of text description. Previous method fails to locate them and is confused with similar keypoints which are appeared during training.

\section{Conclusion and Discussion}
\label{sec:conclusion}

In this paper we introduce the first LLM-based keypoint localization model named LocLLM. Different from previous work that encodes keypoint priors into the model architecture and implicitly learn keypoint relationship from training data, LocLLM explicitly encodes keypoint type and relationship through language description, and utilizes the powerful reasoning capability of LLM to locate keypoints. We conduct experiments on different localization tasks to show the superior generalization ability of LocLLM. As shown in experiments, LocLLM performs well in detecting keypoints from unseen human pose, and locating novel type of keypoints unseen during training. We hope this work inspire future research on generalizable keypoint localization.

Our method can be improved in several aspects. First, the effectiveness of LocLLM relies on accurate textual descriptions. The effectiveness of LocLLM on keypoints that are hard to be described in language remain to be explored, such as facial landmark detection. Second, the huge parameters of LLM require considerable GPU resource to process a large batch of images, hence degrades its efficiency.

\noindent \textbf{Acknowledgement} This work is supported in part by Natural Science Foundation of China under Grant No. U20B2052, 61936011, in part by the Okawa Foundation Research Award.

{
    \small
    \bibliographystyle{ieeenat_fullname}
    \bibliography{main}

\begin{thebibliography}{44}
\providecommand{\natexlab}[1]{#1}
\providecommand{\url}[1]{\texttt{#1}}
\expandafter\ifx\csname urlstyle\endcsname\relax
  \providecommand{\doi}[1]{doi: #1}\else
  \providecommand{\doi}{doi: \begingroup \urlstyle{rm}\Url}\fi

\bibitem[Alayrac et~al.(2022)Alayrac, Donahue, Luc, Miech, Barr, Hasson, Lenc, Mensch, Millican, Reynolds, et~al.]{alayrac2022flamingo}
Jean-Baptiste Alayrac, Jeff Donahue, Pauline Luc, Antoine Miech, Iain Barr, Yana Hasson, Karel Lenc, Arthur Mensch, Katherine Millican, Malcolm Reynolds, et~al.
\newblock Flamingo: a visual language model for few-shot learning.
\newblock \emph{NeurIPS}, 35:\penalty0 23716--23736, 2022.

\bibitem[Carreira et~al.(2016)Carreira, Agrawal, Fragkiadaki, and Malik]{carreira2016human}
Joao Carreira, Pulkit Agrawal, Katerina Fragkiadaki, and Jitendra Malik.
\newblock Human pose estimation with iterative error feedback.
\newblock In \emph{CVPR}, 2016.

\bibitem[Chen et~al.(2023)Chen, Zhang, Zeng, Zhang, Zhu, and Zhao]{chen2023shikra}
Keqin Chen, Zhao Zhang, Weili Zeng, Richong Zhang, Feng Zhu, and Rui Zhao.
\newblock Shikra: Unleashing multimodal llm's referential dialogue magic.
\newblock \emph{arXiv preprint arXiv:2306.15195}, 2023.

\bibitem[Chiang et~al.(2023)Chiang, Li, Lin, Sheng, Wu, Zhang, Zheng, Zhuang, Zhuang, Gonzalez, Stoica, and Xing]{vicuna2023}
Wei-Lin Chiang, Zhuohan Li, Zi Lin, Ying Sheng, Zhanghao Wu, Hao Zhang, Lianmin Zheng, Siyuan Zhuang, Yonghao Zhuang, Joseph~E. Gonzalez, Ion Stoica, and Eric~P. Xing.
\newblock Vicuna: An open-source chatbot impressing gpt-4 with 90\%* chatgpt quality, 2023.

\bibitem[Dai et~al.(2023)Dai, Li, Li, Tiong, Zhao, Wang, Li, Fung, and Hoi]{dai2023instructblip}
Wenliang Dai, Junnan Li, Dongxu Li, Anthony Meng~Huat Tiong, Junqi Zhao, Weisheng Wang, Boyang Li, Pascale Fung, and Steven Hoi.
\newblock Instructblip: Towards general-purpose vision-language models with instruction tuning, 2023.

\bibitem[Hu et~al.(2021)Hu, Shen, Wallis, Allen-Zhu, Li, Wang, Wang, and Chen]{hu2021lora}
Edward~J Hu, Yelong Shen, Phillip Wallis, Zeyuan Allen-Zhu, Yuanzhi Li, Shean Wang, Lu Wang, and Weizhu Chen.
\newblock Lora: Low-rank adaptation of large language models.
\newblock \emph{arXiv preprint arXiv:2106.09685}, 2021.

\bibitem[Ionescu et~al.(2013)Ionescu, Papava, Olaru, and Sminchisescu]{ionescu2013human3}
Catalin Ionescu, Dragos Papava, Vlad Olaru, and Cristian Sminchisescu.
\newblock Human3. 6m: Large scale datasets and predictive methods for 3d human sensing in natural environments.
\newblock \emph{IEEE TPAMI}, 2013.

\bibitem[Ju et~al.(2023)Ju, Zeng, Wang, Xu, and Zhang]{ju2023human}
Xuan Ju, Ailing Zeng, Jianan Wang, Qiang Xu, and Lei Zhang.
\newblock Human-art: A versatile human-centric dataset bridging natural and artificial scenes.
\newblock In \emph{CVPR}, 2023.

\bibitem[Kirillov et~al.(2023)Kirillov, Mintun, Ravi, Mao, Rolland, Gustafson, Xiao, Whitehead, Berg, Lo, et~al.]{kirillov2023segment}
Alexander Kirillov, Eric Mintun, Nikhila Ravi, Hanzi Mao, Chloe Rolland, Laura Gustafson, Tete Xiao, Spencer Whitehead, Alexander~C Berg, Wan-Yen Lo, et~al.
\newblock Segment anything.
\newblock \emph{arXiv preprint arXiv:2304.02643}, 2023.

\bibitem[Lai et~al.(2023)Lai, Tian, Chen, Li, Yuan, Liu, and Jia]{lai2023lisa}
Xin Lai, Zhuotao Tian, Yukang Chen, Yanwei Li, Yuhui Yuan, Shu Liu, and Jiaya Jia.
\newblock Lisa: Reasoning segmentation via large language model.
\newblock \emph{arXiv preprint arXiv:2308.00692}, 2023.

\bibitem[Li et~al.(2021{\natexlab{a}})Li, Bian, Zeng, Wang, Pang, Liu, and Lu]{li2021human}
Jiefeng Li, Siyuan Bian, Ailing Zeng, Can Wang, Bo Pang, Wentao Liu, and Cewu Lu.
\newblock Human pose regression with residual log-likelihood estimation.
\newblock In \emph{ICCV}, 2021{\natexlab{a}}.

\bibitem[Li et~al.(2021{\natexlab{b}})Li, Chen, Shi, Lou, Li, and Lu]{li2021localization}
Jiefeng Li, Tong Chen, Ruiqi Shi, Yujing Lou, Yong-Lu Li, and Cewu Lu.
\newblock Localization with sampling-argmax.
\newblock \emph{NeurIPS}, 2021{\natexlab{b}}.

\bibitem[Li et~al.(2023)Li, Li, Savarese, and Hoi]{li2023blip}
Junnan Li, Dongxu Li, Silvio Savarese, and Steven Hoi.
\newblock Blip-2: Bootstrapping language-image pre-training with frozen image encoders and large language models.
\newblock \emph{arXiv preprint arXiv:2301.12597}, 2023.

\bibitem[Li et~al.(2021{\natexlab{c}})Li, Zhang, Wang, Yang, Yang, Xia, and Zhou]{li2021tokenpose}
Yanjie Li, Shoukui Zhang, Zhicheng Wang, Sen Yang, Wankou Yang, Shu-Tao Xia, and Erjin Zhou.
\newblock Tokenpose: Learning keypoint tokens for human pose estimation.
\newblock In \emph{CVPR}, 2021{\natexlab{c}}.

\bibitem[Li et~al.(2022)Li, Yang, Liu, Zhang, Wang, Wang, Yang, and Xia]{li2022simcc}
Yanjie Li, Sen Yang, Peidong Liu, Shoukui Zhang, Yunxiao Wang, Zhicheng Wang, Wankou Yang, and Shu-Tao Xia.
\newblock Simcc: A simple coordinate classification perspective for human pose estimation.
\newblock In \emph{ECCV}, 2022.

\bibitem[Lin et~al.(2014)Lin, Maire, Belongie, Hays, Perona, Ramanan, Doll{\'a}r, and Zitnick]{lin2014microsoft}
Tsung-Yi Lin, Michael Maire, Serge Belongie, James Hays, Pietro Perona, Deva Ramanan, Piotr Doll{\'a}r, and C~Lawrence Zitnick.
\newblock Microsoft coco: Common objects in context.
\newblock In \emph{ECCV}, 2014.

\bibitem[Liu et~al.(2023)Liu, Li, Wu, and Lee]{liu2023visual}
Haotian Liu, Chunyuan Li, Qingyang Wu, and Yong~Jae Lee.
\newblock Visual instruction tuning.
\newblock \emph{arXiv preprint arXiv:2304.08485}, 2023.

\bibitem[Moon et~al.(2019)Moon, Chang, and Lee]{moon2019camera}
Gyeongsik Moon, Ju~Yong Chang, and Kyoung~Mu Lee.
\newblock Camera distance-aware top-down approach for 3d multi-person pose estimation from a single rgb image.
\newblock In \emph{ICCV}, 2019.

\bibitem[Moon et~al.(2023)Moon, Madotto, Lin, Nagarajan, Smith, Jain, Yeh, Murugesan, Heidari, Liu, et~al.]{moon2023anymal}
Seungwhan Moon, Andrea Madotto, Zhaojiang Lin, Tushar Nagarajan, Matt Smith, Shashank Jain, Chun-Fu Yeh, Prakash Murugesan, Peyman Heidari, Yue Liu, et~al.
\newblock Anymal: An efficient and scalable any-modality augmented language model.
\newblock \emph{arXiv preprint arXiv:2309.16058}, 2023.

\bibitem[Newell et~al.(2016)Newell, Yang, and Deng]{newell2016stacked}
Alejandro Newell, Kaiyu Yang, and Jia Deng.
\newblock Stacked hourglass networks for human pose estimation.
\newblock In \emph{ECCV}, 2016.

\bibitem[Oquab et~al.(2023)Oquab, Darcet, Moutakanni, Vo, Szafraniec, Khalidov, Fernandez, Haziza, Massa, El-Nouby, et~al.]{oquab2023dinov2}
Maxime Oquab, Timoth{\'e}e Darcet, Th{\'e}o Moutakanni, Huy Vo, Marc Szafraniec, Vasil Khalidov, Pierre Fernandez, Daniel Haziza, Francisco Massa, Alaaeldin El-Nouby, et~al.
\newblock Dinov2: Learning robust visual features without supervision.
\newblock \emph{arXiv preprint arXiv:2304.07193}, 2023.

\bibitem[Radford et~al.(2021)Radford, Kim, Hallacy, Ramesh, Goh, Agarwal, Sastry, Askell, Mishkin, Clark, et~al.]{radford2021learning}
Alec Radford, Jong~Wook Kim, Chris Hallacy, Aditya Ramesh, Gabriel Goh, Sandhini Agarwal, Girish Sastry, Amanda Askell, Pamela Mishkin, Jack Clark, et~al.
\newblock Learning transferable visual models from natural language supervision.
\newblock In \emph{ICML}, 2021.

\bibitem[Sagonas et~al.(2013)Sagonas, Tzimiropoulos, Zafeiriou, and Pantic]{sagonas2013300}
Christos Sagonas, Georgios Tzimiropoulos, Stefanos Zafeiriou, and Maja Pantic.
\newblock 300 faces in-the-wild challenge: The first facial landmark localization challenge.
\newblock In \emph{ICCVW}, 2013.

\bibitem[Sharma et~al.(2018)Sharma, Ding, Goodman, and Soricut]{sharma2018conceptual}
Piyush Sharma, Nan Ding, Sebastian Goodman, and Radu Soricut.
\newblock Conceptual captions: A cleaned, hypernymed, image alt-text dataset for automatic image captioning.
\newblock In \emph{ACL}, 2018.

\bibitem[Shi et~al.(2022)Shi, Wei, Li, Ren, and Tan]{shi2022end}
Dahu Shi, Xing Wei, Liangqi Li, Ye Ren, and Wenming Tan.
\newblock End-to-end multi-person pose estimation with transformers.
\newblock In \emph{CVPR}, 2022.

\bibitem[Sun et~al.(2019)Sun, Xiao, Liu, and Wang]{sun2019deep}
Ke Sun, Bin Xiao, Dong Liu, and Jingdong Wang.
\newblock Deep high-resolution representation learning for human pose estimation.
\newblock In \emph{CVPR}, 2019.

\bibitem[Sun et~al.(2017)Sun, Shang, Liang, and Wei]{sun2017compositional}
Xiao Sun, Jiaxiang Shang, Shuang Liang, and Yichen Wei.
\newblock Compositional human pose regression.
\newblock In \emph{ICCV}, 2017.

\bibitem[Sun et~al.(2018)Sun, Xiao, Wei, Liang, and Wei]{sun2018integral}
Xiao Sun, Bin Xiao, Fangyin Wei, Shuang Liang, and Yichen Wei.
\newblock Integral human pose regression.
\newblock In \emph{ECCV}, 2018.

\bibitem[Tompson et~al.(2014)Tompson, Jain, LeCun, and Bregler]{tompson2014joint}
Jonathan~J Tompson, Arjun Jain, Yann LeCun, and Christoph Bregler.
\newblock Joint training of a convolutional network and a graphical model for human pose estimation.
\newblock \emph{NeurIPS}, 2014.

\bibitem[Toshev and Szegedy(2014)]{toshev2014deeppose}
Alexander Toshev and Christian Szegedy.
\newblock Deeppose: Human pose estimation via deep neural networks.
\newblock In \emph{CVPR}, 2014.

\bibitem[Touvron et~al.(2023)Touvron, Martin, Stone, Albert, Almahairi, Babaei, Bashlykov, Batra, Bhargava, Bhosale, et~al.]{touvron2023llama}
Hugo Touvron, Louis Martin, Kevin Stone, Peter Albert, Amjad Almahairi, Yasmine Babaei, Nikolay Bashlykov, Soumya Batra, Prajjwal Bhargava, Shruti Bhosale, et~al.
\newblock Llama 2: Open foundation and fine-tuned chat models.
\newblock \emph{arXiv preprint arXiv:2307.09288}, 2023.

\bibitem[Wang and Zhang(2023{\natexlab{a}})]{wang20233d}
Dongkai Wang and Shiliang Zhang.
\newblock 3d human mesh recovery with sequentially global rotation estimation.
\newblock In \emph{CVPR}, 2023{\natexlab{a}}.

\bibitem[Wang and Zhang(2023{\natexlab{b}})]{wang2023contextual}
Dongkai Wang and Shiliang Zhang.
\newblock Contextual instance decoupling for instance-level human analysis.
\newblock \emph{IEEE TPAMI}, 2023{\natexlab{b}}.

\bibitem[Wang et~al.(2023{\natexlab{a}})Wang, Zhang, Wang, Tian, Huang, and Gao]{wang2023humvis}
Dongkai Wang, Shiliang Zhang, Yaowei Wang, Yonghong Tian, Tiejun Huang, and Wen Gao.
\newblock Humvis: Human-centric visual analysis system.
\newblock In \emph{ACM MM}, 2023{\natexlab{a}}.

\bibitem[Wang et~al.(2022)Wang, Yang, Men, Lin, Bai, Li, Ma, Zhou, Zhou, and Yang]{wang2022ofa}
Peng Wang, An Yang, Rui Men, Junyang Lin, Shuai Bai, Zhikang Li, Jianxin Ma, Chang Zhou, Jingren Zhou, and Hongxia Yang.
\newblock Ofa: Unifying architectures, tasks, and modalities through a simple sequence-to-sequence learning framework.
\newblock In \emph{ICML}, 2022.

\bibitem[Wang et~al.(2023{\natexlab{b}})Wang, Chen, Chen, Wu, Zhu, Zeng, Luo, Lu, Zhou, Qiao, et~al.]{wang2023visionllm}
Wenhai Wang, Zhe Chen, Xiaokang Chen, Jiannan Wu, Xizhou Zhu, Gang Zeng, Ping Luo, Tong Lu, Jie Zhou, Yu Qiao, et~al.
\newblock Visionllm: Large language model is also an open-ended decoder for vision-centric tasks.
\newblock \emph{arXiv preprint arXiv:2305.11175}, 2023{\natexlab{b}}.

\bibitem[Wei et~al.(2021)Wei, Bosma, Zhao, Guu, Yu, Lester, Du, Dai, and Le]{wei2021finetuned}
Jason Wei, Maarten Bosma, Vincent~Y Zhao, Kelvin Guu, Adams~Wei Yu, Brian Lester, Nan Du, Andrew~M Dai, and Quoc~V Le.
\newblock Finetuned language models are zero-shot learners.
\newblock \emph{arXiv preprint arXiv:2109.01652}, 2021.

\bibitem[Xiao et~al.(2018)Xiao, Wu, and Wei]{xiao2018simple}
Bin Xiao, Haiping Wu, and Yichen Wei.
\newblock Simple baselines for human pose estimation and tracking.
\newblock In \emph{ECCV}, 2018.

\bibitem[Xu et~al.(2022)Xu, Zhang, Zhang, and Tao]{xu2022vitpose}
Yufei Xu, Jing Zhang, Qiming Zhang, and Dacheng Tao.
\newblock Vitpose: Simple vision transformer baselines for human pose estimation.
\newblock \emph{arXiv preprint arXiv:2204.12484}, 2022.

\bibitem[Xuan et~al.(2023)Xuan, Guo, Yang, and Zhang]{xuan2023pink}
Shiyu Xuan, Qingpei Guo, Ming Yang, and Shiliang Zhang.
\newblock Pink: Unveiling the power of referential comprehension for multi-modal llms.
\newblock \emph{arXiv preprint arXiv:2310.00582}, 2023.

\bibitem[Ye et~al.(2023)Ye, Xu, Xu, Ye, Yan, Zhou, Wang, Hu, Shi, Shi, et~al.]{ye2023mplug}
Qinghao Ye, Haiyang Xu, Guohai Xu, Jiabo Ye, Ming Yan, Yiyang Zhou, Junyang Wang, Anwen Hu, Pengcheng Shi, Yaya Shi, et~al.
\newblock mplug-owl: Modularization empowers large language models with multimodality.
\newblock \emph{arXiv preprint arXiv:2304.14178}, 2023.

\bibitem[Zhang et~al.(2020)Zhang, Zhu, Dai, Ye, and Zhu]{zhang2020distribution}
Feng Zhang, Xiatian Zhu, Hanbin Dai, Mao Ye, and Ce Zhu.
\newblock Distribution-aware coordinate representation for human pose estimation.
\newblock In \emph{CVPR}, 2020.

\bibitem[Zhang et~al.(2023)Zhang, Wang, Chen, Xu, Zhang, and Tao]{zhang2023clamp}
Xu Zhang, Wen Wang, Zhe Chen, Yufei Xu, Jing Zhang, and Dacheng Tao.
\newblock Clamp: Prompt-based contrastive learning for connecting language and animal pose.
\newblock In \emph{CVPR}, 2023.

\bibitem[Zhu et~al.(2023)Zhu, Chen, Shen, Li, and Elhoseiny]{zhu2023minigpt}
Deyao Zhu, Jun Chen, Xiaoqian Shen, Xiang Li, and Mohamed Elhoseiny.
\newblock Minigpt-4: Enhancing vision-language understanding with advanced large language models.
\newblock \emph{arXiv preprint arXiv:2304.10592}, 2023.

\end{thebibliography}
}

\clearpage
\appendix

\onecolumn
{\centering
\Large
\textbf{LocLLM: Exploiting  Generalizable Human Keypoint Localization \\ via Large Language Model}\\
\vspace{0.5em}
Supplementary Material \\
}

\section{Keypoint Location Description}
In supplemental material we provide the detailed keypoint location description used by LocLLM. Table~\ref{tab:kptdes} shows the all keypoint description, which is generated by ChatGPT and we manually check them with wikipedia to ensure correctness.

\begin{table}[!htb]
\caption{\textbf{The keypoint location description.} These descriptions are generated by ChatGPT and we manually check it with wikipedia.}
\label{tab:kptdes}
\begin{tabular}{@{}p{\linewidth}@{}} 
\toprule
\begin{itemize}
\item \emph{nose}: The \emph{nose} is the central, protruding feature on their face, located just above the upper lip.
\item \emph{left eye}: The \emph{left eye} is the visual organ on the left side of their face, typically located above the left cheek and beside the nose.
\item \emph{right eye}: The \emph{right eye} is the visual organ on the right side of their face, typically located above the right cheek and beside the nose.
\item \emph{left ear}: The \emph{left ear} is the auditory organ on the left side of their head, typically located to the side of the left temple.
\item \emph{right ear}: The \emph{right ear} is the auditory organ on the right side of their head, typically located to the side of the right temple.
\item \emph{left shoulder}: The \emph{left shoulder} is the joint connecting the left arm and the torso, typically situated on the upper left side of the chest.
\item \emph{right shoulder}: The \emph{right shoulder} is the joint connecting the right arm and the torso, typically situated on the upper right side of the chest.
\item \emph{left elbow}: The \emph{left elbow} is the joint connecting the left upper arm and the left forearm, typically situated in the middle of the left arm, between left shoulder and left wrist.
\item \emph{right elbow}: The \emph{right elbow} is the joint connecting the right upper arm and the right forearm, typically situated in the middle of the right arm, between right shoulder and right wrist.
\item \emph{left wrist}: The \emph{left wrist} is the joint connecting the left forearm and the left hand, typically located at the base of the left hand.
\item \emph{right wrist}: The \emph{right wrist} is the joint connecting the right forearm and the right hand, typically located at the base of the right hand.
\item \emph{left hip}: The \emph{left hip} is the joint connecting the left thigh to the pelvis, typically located on the left side of the lower torso.
\item \emph{right hip}: The \emph{right hip} is the joint connecting the right thigh to the pelvis, typically located on the right side of the lower torso.
\item \emph{left knee}: The \emph{left knee} is the joint connecting the left thigh and the left lower leg, typically situated in the middle of the left leg, it is located between the left hip and left ankle.
\item \emph{right knee}: The \emph{right knee} is the joint connecting the upper leg and lower leg on the right side, it is located between the right hip and right ankle.
\item \emph{left ankle}: The \emph{left ankle} is the joint connecting the left lower leg and the left foot, typically located at the base of the left leg.
\item \emph{right ankle}: The \emph{right ankle} is the joint connecting the right lower leg and the right foot, typically located at the base of the right leg.
\item \emph{neck}: The \emph{neck} is the part of the body connecting the head to the torso, typically situated between the shoulders.
\item \emph{pelvis}: The \emph{pelvis} is the bony structure that forms the base of the spine and connects the torso to the lower body, typically located between the left hip and right hip.
\end{itemize} \\ \bottomrule
\end{tabular}
\end{table}

\section{Details of Inference Procedure}
LocLLM is designed to locate only one type of keypoint in a single forward, and a batch inference is needed to locate multiple type of keypoints (\emph{e.g.}, stacking nose, knee \emph{etc} inputs at batch dimension) for an image. This single keypoint inference design allows us to flexibly locate novel type of keypoint by providing corresponding novel descriptions. How to accelerate the inference process is a valuable direction for future research. 

\section{Analysis on Numeric Value Prediction with Cross Entropy Loss}

\textit{Background: How does the Cross Entropy loss handle textual predictions that are slightly off in numeric value, but significantly off in 'text'? For example, 0.399 vs. 0.401. It's not too clear how LocLLM handles this scenario.}

Cross Entropy (CE) loss is the standard training loss for existing autoregressive LLMs, \emph{e.g.}, Llama~\cite{touvron2023llama}, Vicuna~\cite{vicuna2023}, LLaVA~\cite{liu2023visual} and Mini-GPT4~\cite{zhu2023minigpt}. Therefore, it is natural to adopt it to train our LocLLM and align the objective function with pretrained LLMs. Adopting CE loss also allows us to utilize the pretrained classifier in LLM, making it more effective for following visual instruction tuning.

To investigate the capability of handling boundary cases in LocLLM, we count the predictions of LocLLM on keypoints with ground truth in $[0.395, 0.405]$ on COCO \texttt{val} set (with 111 samples in total). All predictions of above samples are in range $[0.361, 0.434]$.

\begin{table}[h]
\setlength{\tabcolsep}{2.3mm}{
\footnotesize
\begin{center}
  \begin{tabular}{l|c|c}
    \toprule
     & Gt$\in [0.395, 0.400)$ & Gt$\in [0.400, 0.405]$\\
    \midrule
    Pred$\in [0.361,0.434]$ & 49 & 62\\
    \midrule
    Pred$\in [0.361,0.399]$ & 29 (59\%) & 25 (40\%)\\
    Pred$\in [0.400, 0.434]$ & 20 (41\%) & 37 (60\%)\\
    \bottomrule
  \end{tabular}
\end{center}
}
\end{table}

It can be observed that for keypoints in $[0.395, 0.399]$( 2nd column), LocLLM can output coordinates both start from 0.3xx (59\%) and 0.4xx (41\%), indicating LocLLM with CE is not biased to one direction in boundary cases. We think it is because LLM is not only focusing on a single token but can understand the context of a set of tokens, making it could handle boundary cases of numeric value well. This contextual understanding capability of LLM is also verified in many NLP and vision-language tasks.

\end{document}